\documentclass{article}
\usepackage{microtype}
\usepackage{graphicx}
\usepackage{booktabs}
\usepackage{hyperref}

\usepackage[accepted]{icml2023}
\usepackage{amsmath}
\usepackage{amssymb}
\usepackage{mathtools}
\usepackage{amsthm}
\usepackage[capitalize,noabbrev]{cleveref}

\usepackage{enumitem}
\usepackage{subcaption}
\usepackage{graphicx}
\usepackage{soul}
\usepackage{wrapfig}
\usepackage{multirow}
\usepackage{color, colortbl}
\usepackage{pifont}
\usepackage{adjustbox}
\theoremstyle{plain}

\theoremstyle{definition}

\theoremstyle{remark}

\definecolor{Gray}{gray}{0.9}
\newcommand*\colourcheck[1]{%
  \expandafter\newcommand\csname #1check\endcsname{\textcolor{#1}{\ding{55}}}%
}
\colourcheck{lightgray}

\newcommand{\xmark}{\textcolor{lightgray}{\ding{55}}}

\DeclareMathOperator*{\argmax}{arg\,max}

\newcommand{\midsepremove}{\aboverulesep = 0mm \belowrulesep = 0mm} 
\newcommand{\midsepdefault}{\aboverulesep = 0.605mm \belowrulesep = 0.984mm}
\usepackage[textwidth=14mm,textsize=tiny]{todonotes}

\icmltitlerunning{Learning Representations without Compositional Assumptions}

\begin{document}

\twocolumn[
\icmltitle{Learning Representations without Compositional Assumptions}
\icmlsetsymbol{equal}{*}

\begin{icmlauthorlist}
\icmlauthor{Tennison Liu}{cam}
\icmlauthor{Jeroen Berrevoets}{cam}
\icmlauthor{Zhaozhi Qian}{cam}
\icmlauthor{Mihaela van der Schaar}{cam,ati}
\end{icmlauthorlist}

\icmlaffiliation{cam}{DAMTP, University of Cambridge, Cambridge, UK}
\icmlaffiliation{ati}{Alan Turing Institute, London, UK}

\icmlcorrespondingauthor{Tennison Liu}{tl522@cam.ac.uk}
\icmlkeywords{Multi-view Learning}

\vskip 0.3in
]
\printAffiliationsAndNotice{} 

\begin{abstract}
This paper addresses unsupervised representation learning on tabular data containing multiple views generated by distinct sources of measurement. Traditional methods, which tackle this problem using the multi-view framework, are constrained by predefined assumptions that assume feature sets share the same information and representations should learn globally shared factors. However, this assumption is not always valid for real-world tabular datasets with complex dependencies between feature sets, resulting in localized information that is harder to learn. To overcome this limitation, we propose a data-driven approach that learns feature set dependencies by representing feature sets as graph nodes and their relationships as learnable edges. Furthermore, we introduce $\texttt{LEGATO}$, a novel hierarchical graph autoencoder that learns a smaller, latent graph to aggregate information from multiple views dynamically. This approach results in latent graph components that specialize in capturing localized information from different regions of the input, leading to superior downstream performance.

\end{abstract}

\section{Introduction} \label{sec:introduction}

Tabular datasets encountered in the real world often contain distinct feature sets, or views, that originate from different sources of measurement. For instance, the UK Biobank \citep{bycroft2018uk} contains measurements of sociodemographic factors, heart and lung function, genomic data, and electronic health records, each providing information on a different aspect of a patient's medical state, but also dependent on one another to form a holistic medical context.

While different feature sets can be consolidated into a single table, doing so can result in suboptimal learning performance due to heterogeneity among feature sets and the loss of valuable relational information. A common approach is then \emph{multi-view learning} \citep{xu2013survey}, which examines each feature set separately and integrates information from multiple views to learn representations. This task can be difficult, particularly when labels for supervision are not available, which can help disambiguate the dependencies between views and task-relevant information. In the unsupervised learning setting, models rely on data assumptions and inductive biases to learn good representations automatically \citep{locatello2019challenging}. 

Existing multi-view learning methods often rely on \emph{compositional assumptions}, which assume that information is distributed and should be aggregated in predetermined patterns. The classic multi-view inductive bias assumes that views provide similar task-relevant information \citep{yan2021deep}, guiding how information is aggregated, with the goal of learning robust and generalized representations that are invariant across views \citep{federici2019learning}. These assumptions have been widely used in image, text, and speech domains, such as audio-visual speech recognition \citep{huang2013audio} and image-caption models \citep{radford2021learning}, where the settings are more controlled and the number of views is limited. In these domains, systematic and aligned data collection ensures maximal information overlap between feature sets, making inter-view relationships and information aggregation strategies known in advance. 

However, these assumptions may not hold for tabular multi-view data, especially those collected \emph{in-the-wild} (ITW), where relationships between feature sets are significantly more opaque. Examples of this include electronic health records \citep{johnson2023mimic}, biobanks \citep{nagai2017overview}, and stock market data \citep{xu2018stock}. In these datasets, information is more likely to exist in localized clusters of views in unknown patterns, rather than being globally present in all views \citep{xu2013survey}. This is particularly true when dealing with tabular problems that typically have more than two feature sets. Our findings indicate that compositional assumptions are inadequate when learning on tabular data collected in-the-wild, failing to capture localized information in representations.

To overcome this challenge, we propose a method to model relationships between feature sets and dynamically aggregate potentially localized information. We represent feature sets as graph nodes and their relationships as learnable edges. Furthermore, we introduce the \underline{L}at\underline{e}nt \underline{G}raph \underline{A}u\underline{TO}encoder (\texttt{LEGATO}), a novel graph neural network that learns a smaller, latent graph. This architecture innovates on existing autoencoder architectures that learn compact node embeddings, but do so on the same topology as the input graph. Our method learns a smaller \emph{graph}, which is crucial, as it allows for end-to-end learning of information aggregation strategies without relying on predefined assumptions. We term the latent graph a \emph{decomposable representation} to emphasize that, by design, it can be decomposed into node representations that specialize in aggregating information from different regions of the input. We evaluate the effectiveness of our method by testing its ability to transfer to downstream tasks, as a good representation should facilitate subsequent problem-solving.

\textbf{Contributions.}~\textbf{1.} We identify the challenges associated with learning representations from heterogeneous tabular feature sets collected in real-world settings and showcase the limitations of existing unsupervised learning methods that heavily rely on predefined compositional assumptions. \textbf{2.} Instead of relying on predefined assumptions, we propose a novel approach that treats feature sets as graphs to capture dependencies, which to the best of our knowledge, is a novel way to represent multi-view data. \textbf{3.} We introduce \texttt{LEGATO}, a novel graph autoencoder architecture that learns a smaller latent graph. This smaller graph induces a decomposable representation by dynamically aggregating localized information in a hierarchical manner. We conduct simulation studies to demonstrate the effectiveness of our model in learning data-driven aggregation strategies. Moreover, we showcase the superior downstream performance of our method on multiple real-world datasets.

\section{Problem Definition} \label{sec:background}
\subsection{Notation} \label{sec:notation}
In this paper, we use the terms ``feature sets'' and ``views'' interchangeably. We consider $K$ different feature sets, depicting one instance $X = \{X^k : k \in [K]\}$. Each $X^k$ is sampled from a space $\mathcal{X}^k \subseteq \mathbb{R}^{d^k}$, and $\mathcal{X} = \mathcal{X}^1 \times \dots \times \mathcal{X}^k$. With $X$ the random variable, we have $x = \{x^k : k \in [K]\}$ as its realization. For each $x^k$, we have a $d$-dimensional view embedding $\smash{h^k \in \mathcal{H}^k \subseteq \mathbb{R}^d}$ produced using an encoder function $\smash{g^k : \mathcal{X}^k \to \mathcal{H}^k}$.\footnote{We assume the embedding dimension is $d$ for all views for notation convenience, but this restriction is not necessary.} Correspondingly, $\smash{f^k: \mathcal{H}^k\rightarrow \mathcal{X}^k}$ denotes the view decoder function. We are agnostic to the exact architecture of $\smash{g^k(\cdot)}$ and $\smash{f^k(\cdot)}$ for generality.  We have access to a dataset $\smash{\mathcal{D} = \{x_i\}_{i=1}^N}$, with $N$ iid samples. We use superscript to indicate view and subscript for the sample, such that $\smash{x_i^k}$ is the $k^{th}$ view of the $i^{th}$ sample. When the context is clear, we may drop the subscript to declutter exposition. 

\subsection{Challenges of Learning In-The-Wild} \label{sec:problem_definition}

\begin{figure}[t!]
     \centering
     \begin{subfigure}[b]{0.48\columnwidth}
         \includegraphics[width=\linewidth]{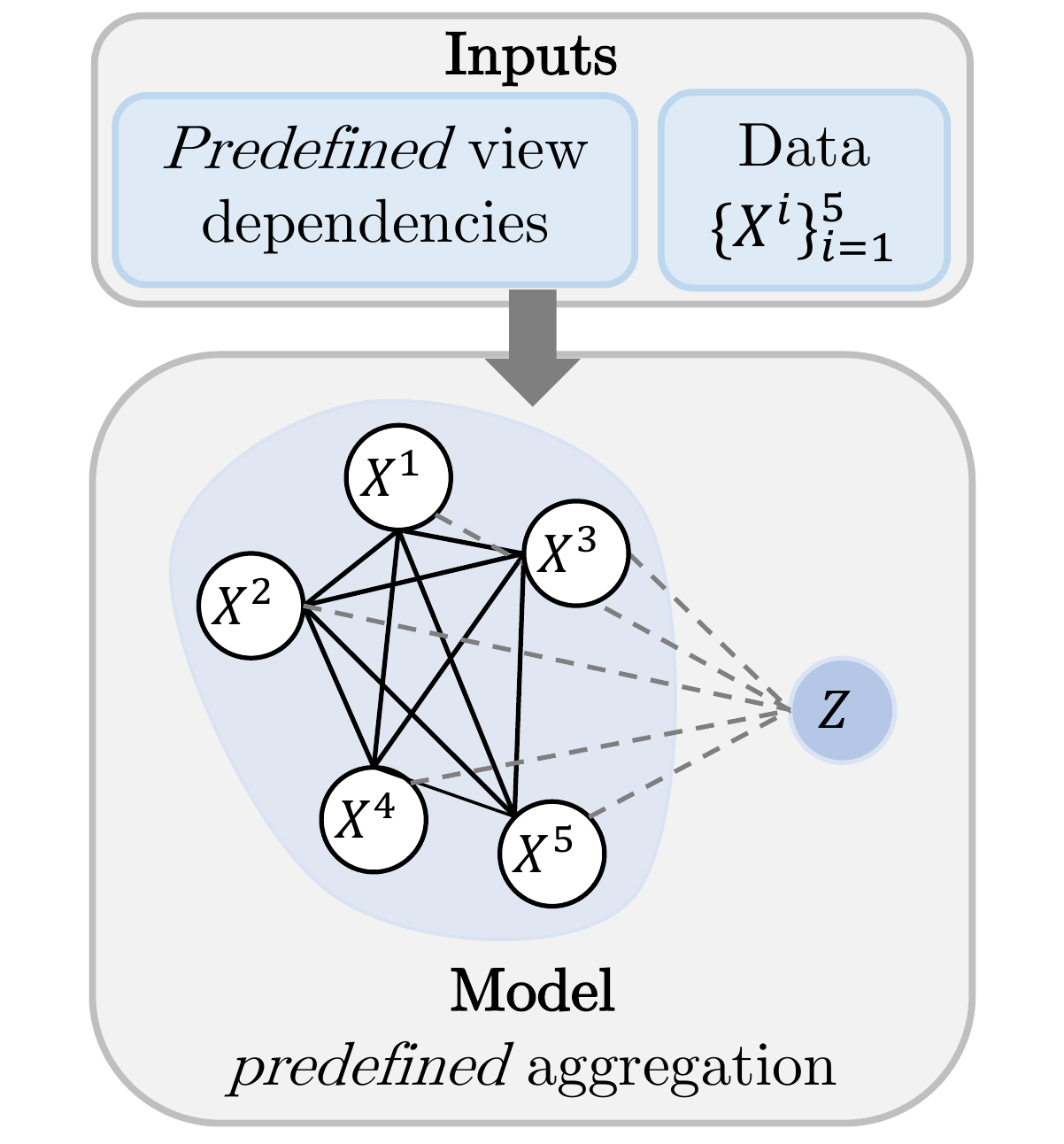}
         \caption{\textit{Predefined} aggregation.}
         \label{fig:global_assumption}
     \end{subfigure}
     \hfill
     \begin{subfigure}[b]{0.48\columnwidth}
         \includegraphics[width=\linewidth]{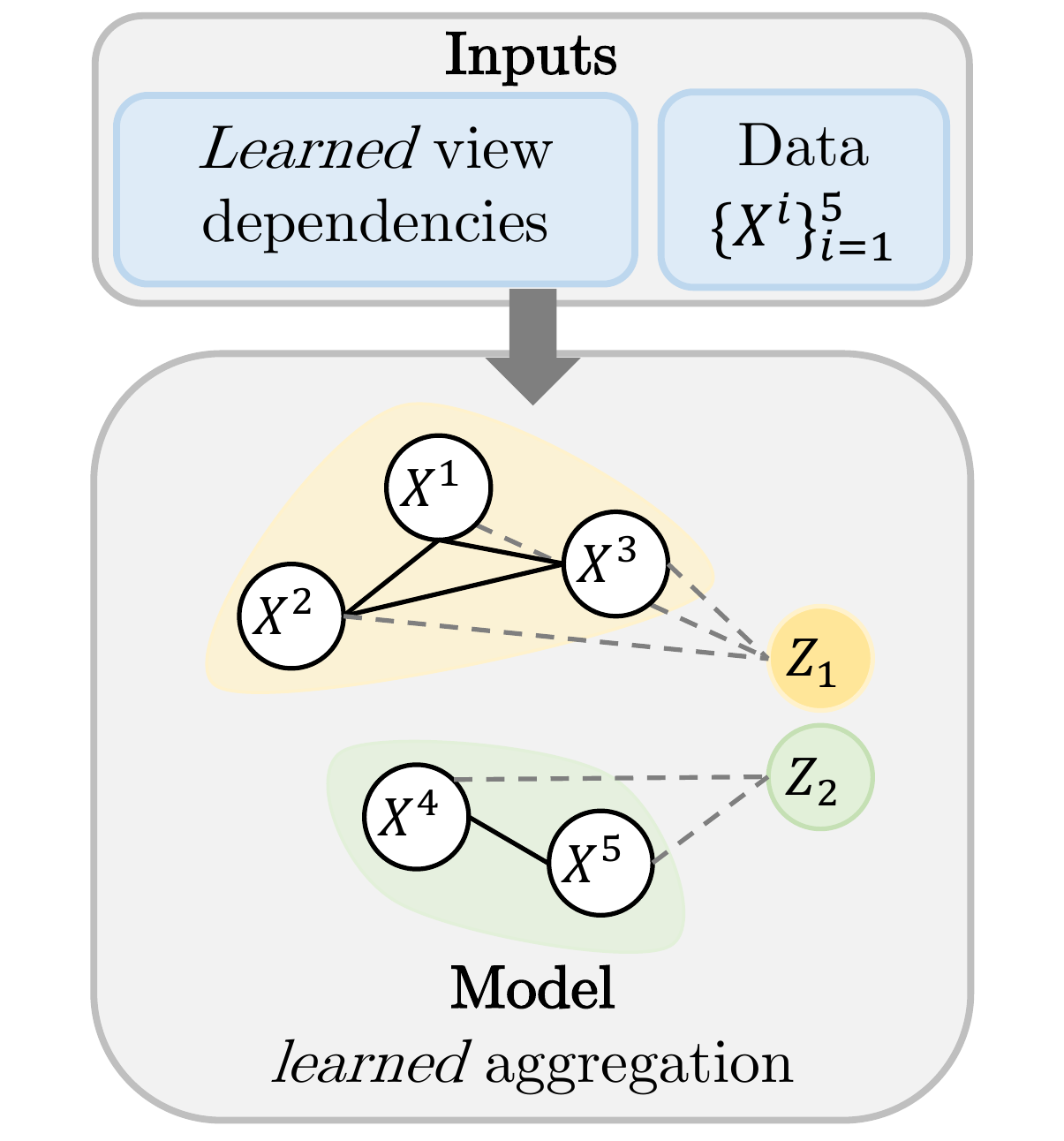}
         \caption{\textit{Learned} aggregation.}
         \label{fig:local_assumption}
     \end{subfigure}
\caption{\textbf{Dynamically aggregating information}. Solid lines represent information sharing and dashed lines represent aggregation. Existing methods (\ref{fig:global_assumption}) assume views share the same information and aggregate information globally. In comparison, our method (\ref{fig:local_assumption}) learns dependencies and  aggregation strategy in a data-driven manner.}
\label{fig:global_vs_local}
\end{figure}

\begin{figure}[t!]
     \centering
     \begin{subfigure}[b]{0.99\columnwidth}
         \includegraphics[width=\linewidth]{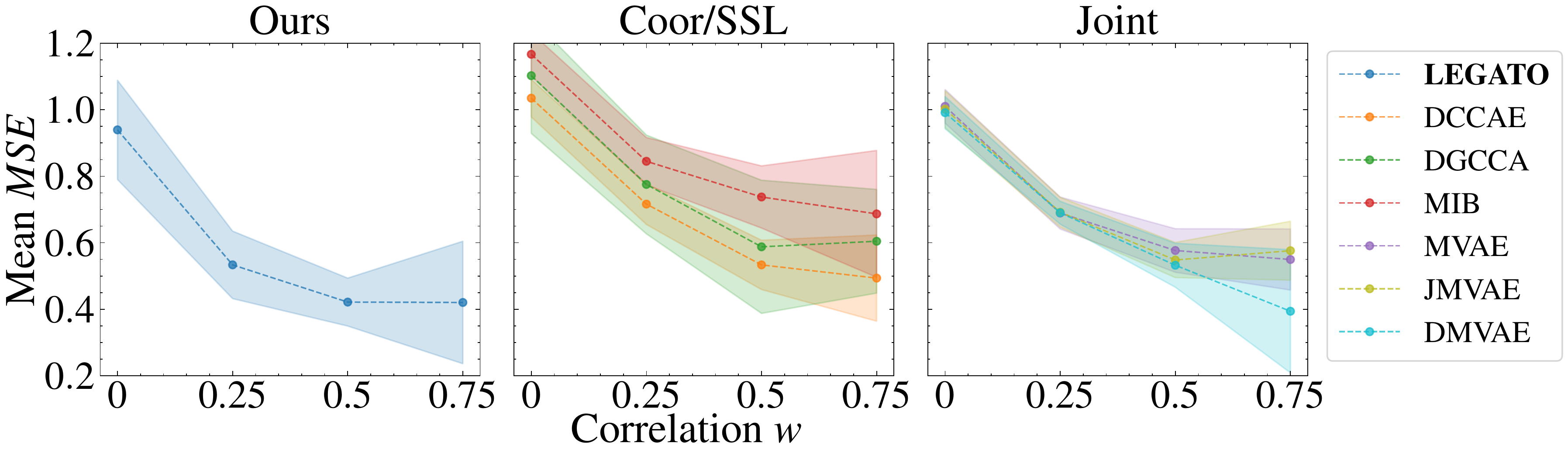}
         \caption{Views are \emph{globally} correlated.}
         \label{fig:global_corr}
     \end{subfigure}
     \begin{subfigure}[b]{0.99\columnwidth}
         \includegraphics[width=\linewidth]{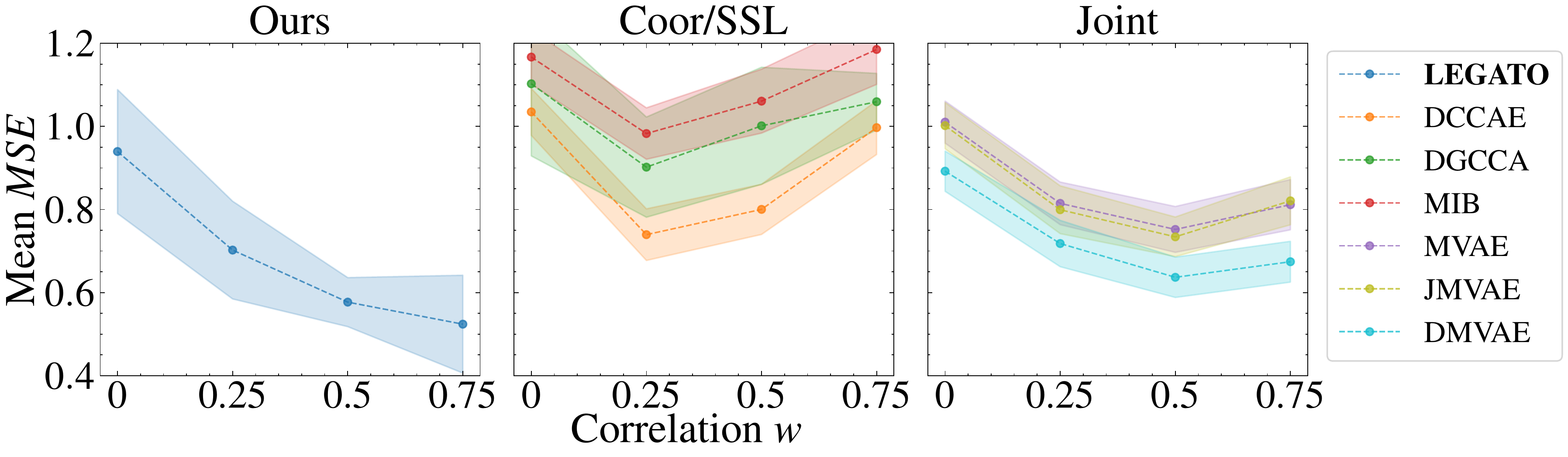}
         \caption{Views are \emph{locally} correlated. }
         \label{fig:local_corr}
     \end{subfigure}
\caption{\textbf{Effect of view correlation on learning ($\mathbf{K}$=$\mathbf{6}$).} When views are globally correlated, higher correlation improves performance for all models. When local correlation increases, the performance of existing methods deteriorates as they fail to learn localized information.}
\label{fig:failure_mode}
\vspace{-1em}
\end{figure}

\textbf{Compositional assumptions.} 
Compositional assumptions are two-fold: they reflect beliefs on how information is shared between feature sets, and how information should be aggregated in a representation. The multi-view assumption is the predominant compositional assumption made in existing works---it posits that important information co-occurs in all available views, leading to a focus on maximizing mutual information among multiple view representations \citep{federici2019learning}. This approach has been successfully applied in many domains, especially image, text, and speech, where it is known apriori (e.g. through careful data collection) that the semantically meaningful variations exist in all signals \citep{vrigkas2015review,radford2021learning}. By learning shared information, these methods improve the robustness and generalization of multi-view representations. More recently, methods have also considered the possibility that each view may contain unique information not present in other views \citep{xu2013survey}, with the aim of retaining both view-specific and globally shared information.

\textbf{Multi-view data collected in-the-wild.} 
Tabular feature sets collected in-the-wild (ITW) present a different challenge, as information is rarely presented in known patterns across different views. We argue that tabular multi-view data found in the real world are characterized by two main features: $\blacktriangleright$ \textbf{Localized information} - where different sources of information are concentrated in localized subsets of views, as opposed to the globally shared information assumed by existing methods, and $\blacktriangleright$ \textbf{A larger number of views} - resulting in more complex dependencies and localized clusters of information. We provide further discussions and detailed case studies on these feature sets and their characteristics in \Cref{app:itw_data}.

These characteristics make the representation learning task more challenging. Existing methods use the multi-view inductive bias to infer a common representation $z$ (and optionally a set of view-specific representations $\smash{\{z_i\}_{i=1}^K}$), leading to a global aggregation of information, as visualized in \Cref{fig:global_assumption}. However, these assumptions are inadequate to address problems ITW, which contain localized information that manifests in unknown ways. Additionally, the large number of possible view combinations (combinatorial in $K$) makes it infeasible to explicitly consider different local aggregation patterns. Our method, as depicted in \Cref{fig:local_assumption}, addresses this challenge by proposing a novel approach to dynamically learn dependencies and aggregate information, without relying on predefined assumptions.

\textbf{Learning challenges.} Given the learning capacity and expressiveness of modern neural networks, it is natural to wonder whether incorrectly specified compositional assumptions are truly detrimental in practice. While representations may be biased, they can still implicitly learn all localized sources of information. We empirically show that this is not the case in a simulation study (described in \Cref{sec:simulation}), where the downstream task is to predict the latent variables that generated each view. Existing methods perform better as the global correlation between latent variables increases (\Cref{fig:global_corr}). This is intuitive because views contain more information about latent variables in other views, which can be effectively learned using the multi-view inductive bias. However, when latent variables are only locally correlated (\Cref{fig:local_corr}), increased correlation does not lead to improved performance. This is because higher correlation only provides locally useful information, which is overlooked when incorrect compositional assumptions are used. 

\begin{figure*}[t!]
\centering
\includegraphics[width=0.99\textwidth]{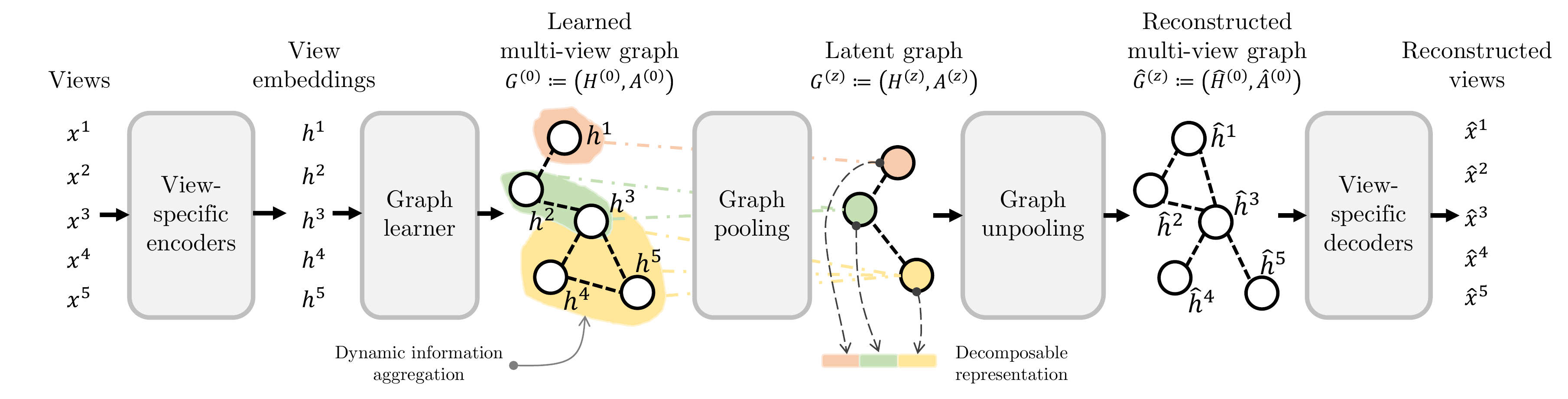}
\caption{\textbf{High-level illustration of LEGATO.} The latent graph dynamically pools information by considering both view embeddings and dependencies. The latent graph returns a decomposable representation for downstream tasks.}
\label{fig:mlg_learning_arch}
\end{figure*}

\section{Proposed Method} \label{sec:method}
We propose a framework for \emph{learning} information aggregation patterns from data without predefined compositional assumptions. This requires accounting for localized information sharing between views, which can be naturally represented using graphs. Our method makes two contributions: first, we learn the view dependencies as edges in a graph, which, to the best of our knowledge, is a novel way to represent multi-view data. Second, we introduce \texttt{LEGATO}, a novel graph autoencoder that learns a smaller latent graph. The latent graph produces a decomposable representation that aggregates localized information. To complete the autoencoder, the latent graph is unpooled to reconstruct each view individually, with the hierarchical process trained end-to-end. Our proposed method is illustrated in \Cref{fig:mlg_learning_arch}.

\subsection{Learning the Multi-view Graph} \label{sec:attention_mechanism}
We define an initial graph on view embeddings, where nodes represent views and edges represent the inter-view relationships, i.e. $\smash{G^{(0)} \coloneqq \left(H^{(0)}, A^{(0)}\right)}$.  $\smash{A^{(0)} \in [0, 1]^{K\times K}}$ is the adjacency matrix between $K$ views and $\smash{H^{(0)} \in \mathbb{R}^{K \times d}}$ is the node feature matrix, where the $k^{th}$ row is the view embedding $h^k$. We are agnostic to the view encoder-decoder architecture and first obtain view embeddings $\smash{h^k=g^k(x^k)}$ independently for each view $k \in [K]$. 

The adjacency matrix $A^{(0)}$ is rarely known. In the most general setting, every node can be connected to every other node, ignoring localized structure. This reflects the multi-view inductive bias, which assumes that each view shares the same information with all other views (as shown in \Cref{fig:global_assumption}). Clearly, this is not the case ITW, as certain views will only share information locally with other views. 

We propose to learn the localized graph structure. Specifically, $\texttt{GRAPHLEARNER}:\mathbb{R}^{K\times d}\rightarrow [0, 1]^{K\times K}$, which takes as input the view embeddings and returns the adjacency matrix. We first apply a non-linear transformation to each view embedding:
\begin{equation}
    e_i = \texttt{LeakyReLU}\left(W[h_i\Vert{1}_i]\right)
\end{equation}
where $\smash{{W}\in\mathbb{R}^{d'\times f}}$ applies a linear transformation, followed by a \texttt{LeakyReLU}$(\cdot)$ activation. We encode view information for view $i$ through the concatenation operation $\Vert$ of $h_i$ and the one-hot encoding ${1}_i$ to obtain a $d'$-dimensional input. Then, we compute the inner product between views, normalized by the \texttt{sigmoid} function $\sigma(\cdot)$ :
\begin{equation}
    A_{ij}^{(0)} = \sigma(e_i^T\cdot e_j)
\end{equation}

The normalized coefficients take values $\in [0, 1]$ to represent the dependence between views. Note that the mechanism is invariant to the ordering of inputs and that $\smash{A^{(0)}}$ is a symmetric matrix. We additionally apply a threshold function to $\smash{A^{(0)}}$, where entries $<0.1$ are considered uninformative and zeroed out. As we want informative local neighbors to be found, we add a regularization term $\smash{\mathcal{L}_{spar}=\frac{1}{NK^2}\sum_{i=1}^N\lVert A^{(0)}_i\rVert_1}$, where $\smash{\lVert\cdot\rVert}_1$ denotes the $p=1$ matrix norm. This term encourages sparsity in the adjacency matrix and reduces the learning of spurious dependencies between views (e.g. by learning a fully-connected graph).

\textbf{A distinction.} We emphasize that our goal is not \emph{relational inference}, which seeks to infer relationships between views from observation data \citep{kipf2018neural,hajiramezanali2020bayrel,hasanzadeh2021morel}. In this problem, a correctly recovered relational structure \emph{is} the object of inference. This stands in stark contrast to our work, where a partially correct structure is satisfactory, as our main purpose is to aggregate information while considering local dependencies. As we shall show later, even learning a partially correct structure can greatly improve the learned representations.

\subsection{LEGATO: Latent Graph Autoencoder}
After learning an initial adjacency matrix, the next step is to aggregate information shared between views in a latent graph. To do this, we leverage the intuition that views with similar information should be aggregated together. We introduce \texttt{LEGATO}, a hierarchical procedure that learns a latent graph while pooling essential information \citep{cai2021graph}. In more detail, we transform  $\smash{G^{(0)}}$ through a pooling step to obtain a latent graph $\smash{G^{(z)}}$. This transformation pools information shared between views, so that each latent node aggregates localized information. Next, we take an unpooling step to reconstruct the graph $\smash{\hat{G}^{(0)}}$, and the entire hierarchical model is trained end-to-end as an autoencoder. 

We use \emph{graph neural networks} (GNN) to learn the latent graph representation \citep{gilmer2017neural,zhou2020graph}. However, existing graph autoencoders are unsuitable for our purposes. The latent graphs in existing works learn compact node embeddings on the same graphical structure as the input graph, where similarity objectives are used to encourage embeddings of topologically connected nodes to be more similar \citep{kipf2016variational,simonovsky2018graphvae}. In contrast, the latent graph learned in \texttt{LEGATO} is a smaller, pooled graph that aggregates information from input views with stronger dependencies. We provide an overview of GNN methods and elaborate on related graph autoencoder methods in \Cref{app:gnn}.

\textbf{Graph pooling.}~The latent graph $\smash{G^{(z)}\coloneqq (H^{(z)}, A^{(z)})}$ is a pooled graph with $K'<K$ nodes. Here, $\smash{A^{(z)}\in [0, 1]^{K'\times K'}}$ and $\smash{H^{(z)}\in \mathbb{R}^{K'\times r}}$, where each row is a $r$-dimensional latent node embedding. We propose a graph pooling operation $\smash{(H^{(z)}, A^{(z)}) = \text{POOL}(H^{(0)}, A^{(0)})}$ by adapting the \texttt{DiffPool} algorithm \citep{ying2018hierarchical}. In our experiments, we set $K'=K/2$, which was found to be a robust setting. Additionally, we note that by setting $K'=1$, we can perform global aggregation, similar to existing methods.

\textbf{Pooling strategy.}~The pooling strategy is learned through a separate network that considers localized dependencies and view embeddings. This is different from traditional compositional assumptions that predefine the pattern of aggregation. Specifically, we learn a pooling matrix $\smash{P \in[0, 1]^{K\times K'}}$ in an input-dependent way by considering both view embeddings in $\smash{H^{(0)}}$ and view dependencies in $\smash{A^{(0)}}$. Intuitively, views that are dependent on each other likely contain similar information and should be aggregated together. To operationalize this insight, we learn the pooling matrix through a GNN:
\begin{equation}
    P = \texttt{softmax}\left(\text{GNN}_{pool}(A^{(0)}, H^{(0)})\right)
    \label{eq:assignment}
\end{equation}
The $\texttt{softmax}(\cdot)$ is applied in a \emph{row}-wise fashion. Consequently, $P$ indicates how information should be aggregated, where $\smash{P_{ij}}$ describes the contribution of the $i^{th}$ view in the multi-view graph to the $j^{th}$ node in the latent graph.

\textbf{Latent embeddings.}~We employ a separate GNN to update view embeddings using neighboring views' embeddings through message passing. This network  produces $\smash{Z\in \mathbb{R}^{K\times r}}$, where each row now contains the updated $r$-dimensional embedding for each view:
\begin{equation}
    Z = \text{GNN}_{embed}(A^{(0)}, H^{(0)})
    \label{eq:transform}
\end{equation}
By combining the pooling matrix and the transformed embeddings in \Cref{eq:assignment,eq:transform}, we can now define the complete $\texttt{POOL}$ operation. Mathematically, we can obtain the latent graph using the following equations:
\begin{align}
    A^{(z)} &= P^TA^{(0)}P \in \mathbb{R}^{K'\times K'} \label{eq:refinement_a} \\
    H^{(z)} &= P^TZ \in \mathbb{R}^{K'\times r} \label{eq:refinement_x}
\end{align}
As in \Cref{eq:refinement_x}, the latent embeddings are constructed through a weighted combination of transformed view embeddings using the pooling strategy in $P$. This reflects the intuition that if a latent node pools information from a set of views, then its embedding should be constructed from those views. Correspondingly, the latent adjacency matrix \Cref{eq:refinement_a} considers existing connectivity strength in $\smash{A^{(0)}}$ and $P$ to compute a weighted sum of edges between neighboring nodes.

\textbf{Orthogonality loss.} In practice, it can be difficult to train the pooling function $\text{GNN}_{pool}$ using only gradient signal from an unsupervised loss. Instinctively, the function can learn a degenerate assignment where information is evenly pooled in the latent nodes, akin to the degeneracy of clustering \citep{alguwaizani2012degeneracy}. This would achieve the opposite of our desired objective, as we want latent nodes to aggregate different localized information. To alleviate this issue, we introduce an orthogonality regularization:
\begin{equation}
    \mathcal{L}_{orth}= \frac{1}{N}\sum_{i=1}^N\frac{1}{C}\sum_{k=2}^{K'}\sum_{j=1}^{k-1}\left\lVert\rho\left(h^k_i, h^j_i\right)\right\rVert_1
\end{equation}
where $\smash{C=\frac{K'\cdot(K'-1)}{2}}$ is the number of pairwise correlations and $\rho(\cdot, \cdot)$ is calculated using cosine similarity. This term encourages orthogonality in the embeddings by de-correlating them. This encodes the intuition of decomposable representations, that each component should specialize in aggregating information from different local regions of the input, resulting in better representations for downstream tasks \citep{mathieu2019disentangling}.

\subsection{Completing the Graph Autoencoder}
\textbf{Graph unpooling.} The unpooling step decodes the original multi-view input from the pooled latent graph. We define the unpooling step $\smash{(\hat{H}^{(0)}, \hat{A}^{(0)})=\texttt{UNPOOL}(A^{(z)}, H^{(z)})}$, where, $\smash{\hat{H}^{(0)}}$ and $\smash{\hat{A}^{(0)}}$ have the same dimensions as the input multi-view graph. Unpooling is mathematically identical to the pooling steps described in \Cref{eq:transform,eq:assignment,eq:refinement_a,eq:refinement_x}. The intuition is also similar, in that the input graph is reconstructed based on a weighted combination of adjacency patterns and embeddings of the latent nodes. After the unpooling step, the node embeddings are passed through the corresponding view-specific decoders to reconstruct the views $\smash{\hat{x} = \{\hat{x}^k\: :\: k \in [K]\}}$.

\textbf{Training.}~It is worth mentioning that multiple pooling and unpooling steps can be stacked, leading to the network gradually operating on more compressed latent graphs. For training the hierarchical model, we specify a reconstruction loss defined on the multi-view graphs:
\begin{equation}
\begin{split}
\mathcal{L}_{recon} = &\frac{1}{NK}\sum_{i=1}^N\sum_{k=1}^K\lVert x^k_i-\hat{x}^k_i \rVert^2_2 \\&+ \frac{1}{N}\sum_{i=1}^N\lVert A^{(0)}-\hat{A}^{(0)}\rVert^2_2
\end{split}
\end{equation}
where the first term is a loss on reconstructed node embeddings and the second term is a loss on the recovered graph structure, together forming the graph reconstruction loss. This loss is combined with the regularization terms to form the training objective: $\smash{\mathcal{L}_{recon}+\alpha\mathcal{L}_{orth}+\beta\mathcal{L}_{spar}}$, where $\mathcal{L}_{spar}$
 regularizes the sparsity of the learned multi-view graph to reduce learning of spurious dependencies between views and $\mathcal{L}_{orth}$
 is an orthogonality regularization that decorrelates latent node embeddings. $\alpha$ and $\beta$ are the corresponding weighting terms for the two regularization terms. This expression and the hierarchical procedure are fully differentiable and can be trained end-to-end using auto-grad techniques.

\subsection{Latent Graph and Decomposable Representations}
Existing unsupervised algorithms learn a latent representation that integrates different sources of information shared between views. However, this often results in representations that entangle localized information and are difficult to differentiate for downstream models. Our learned latent graph is decomposable and is expected to better preserve information and make it more amenable for downstream tasks  \citep{lipton2018mythos}.

\textbf{Decomposable representations.}~We claim that the latent graph is decomposable, as nodes act as specialized components that extract localized information from different regions in the input, and are encouraged to be orthogonal through  $\mathcal{L}_{orth}$. To make the representation more suitable for downstream models, we include an additional readout step that converts the latent graph into a vector $\smash{\texttt{READOUT}:\mathbb{R}^{K'\times r}\times[0, 1]^{K'\times K'}\rightarrow \mathbb{R}^r}$. We use mean pooling to aggregate the node embeddings $\smash{z = \frac{1}{K'}\sum_{k=1}^{K'}h^k}$ and produce a vector representation that is composed of orthogonal components. Future works can consider more advanced readout strategies, including those that take into account graph topology \citep{buterez2022graph}. 

Our approach can be informally compared to convolutional networks that extract localized information from natural images, which contain features in localized patches \citep{lecun2010convolutional}. Importantly, pixels are related in a grid-like pattern and convolutional networks exploit this structure to learn and pool localized information. In our case, the relationships between views are not known a priori. Instead of making predefined assumptions, we model multi-view data as graphs and learn localized dependencies as edge weights. Subsequently, our graph autoencoder facilitates locality in information aggregation to compose representations.

\section{Related Works} \label{sec:related_works}
This work proposes a novel graph autoencoder for unsupervised representation learning on tabular multi-view data collected ITW. As such, there are two lines of related works: multi-view learning methods and GNN architectures.

\begin{figure}[t!]
\centering
\captionof{table}{\textbf{Related works.} Comparison of representative \emph{unsupervised multi-view learning methods} based on \textbf{training objective}, \textbf{assumed generative view (Asm)}, and desiderata: \textbf{(1)} scales to $>2$ views, \textbf{(2)} learns localized information, and \textbf{(3)} dynamically learns aggregation strategy.}
\label{tab:related_works}
\begin{adjustbox}{max width=0.48\textwidth}        
\midsepremove
% \small
\begin{tabular}{c|l|c|c|ccc|}
\cmidrule{2-7}
 & \textbf{Method} & \textbf{Objective} & \textbf{Asm} & \textbf{(1)} & \textbf{(2)} & \textbf{(3)} \\ \midrule
\multicolumn{1}{|c|}{} & \citet{suzuki2016joint} & Recon & fig \ref{fig:consensus_principle} & \checkmark  & \xmark & \xmark \\\cmidrule{2-7}
\multicolumn{1}{|c|}{} & \citet{wu2018multimodal} & Recon & fig \ref{fig:consensus_principle} & \checkmark & \xmark & \xmark \\\cmidrule{2-7}
 \multicolumn{1}{|c|}{} & \citet{zhang2019ae2} & Recon & fig \ref{fig:complementary_principle} & \checkmark  & \checkmark & \xmark \\\cmidrule{2-7}
\multicolumn{1}{|c|}{\parbox[t]{2mm}{\multirow{-4}{*}{\rotatebox[origin=c]{90}{\emph{Joint}}}}} & \citet{lee2021private} & Recon & fig \ref{fig:complementary_principle} & \checkmark  & \checkmark & \xmark \\ \midrule
 \multicolumn{1}{|c|}{}& \citet{andrew2013deep} & CCA & fig \ref{fig:consensus_principle} & \xmark & \xmark & \xmark \\\cmidrule{2-7}
\multicolumn{1}{|c|}{} & \citet{wang2015deep} & CCA & fig \ref{fig:consensus_principle} & \xmark & \checkmark & \xmark \\\cmidrule{2-7}
\multicolumn{1}{|c|}{\parbox[t]{2mm}{\multirow{-3}{*}{\rotatebox[origin=c]{90}{\emph{Coor.}}}}} & \citet{benton2017deep} & CCA & fig \ref{fig:consensus_principle} & \checkmark & \xmark & \xmark \\ \midrule
\multicolumn{1}{|c|}{} & \citet{federici2019learning} & MIB & fig \ref{fig:consensus_principle} & \xmark & \xmark & \xmark \\ \cmidrule{2-7}
\multicolumn{1}{|c|}{} & \citet{radford2021learning} & Contrastive & fig \ref{fig:consensus_principle} & \xmark & \xmark & \xmark \\ \cmidrule{2-7}
\multicolumn{1}{|c|}{\parbox[t]{2mm}{\multirow{-3}{*}{\rotatebox[origin=c]{90}{\emph{SSL}}}}} & \citet{tian2020contrastive} & Contrastive & fig \ref{fig:consensus_principle} & \checkmark & \xmark & \xmark \\ \midrule
& \cellcolor[HTML]{C0C0C0}\textbf{LEGATO} & \cellcolor[HTML]{C0C0C0}\textbf{Recon} & \cellcolor[HTML]{C0C0C0}\textbf{NA} & \cellcolor[HTML]{C0C0C0}\checkmark & \cellcolor[HTML]{C0C0C0}\checkmark & \cellcolor[HTML]{C0C0C0}\checkmark \\
\cmidrule{2-7}
\end{tabular}
\midsepdefault
\end{adjustbox} \\

\vspace{1em}
% \refstepcounter{table}
% \refstepcounter{table}
% \refstepcounter{table}
\centering
\begin{subfigure}[b]{0.107\textwidth}
 \centering
 \includegraphics[width=\textwidth]{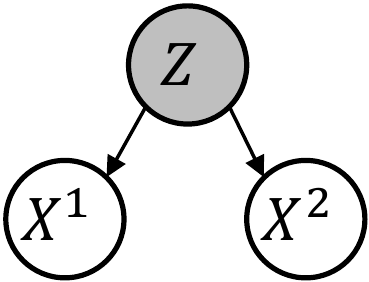}
 \caption{}
 \label{fig:consensus_principle}
\end{subfigure}
\hspace{4em}
\begin{subfigure}[b]{0.15\textwidth}
 \centering
 \includegraphics[width=\textwidth]{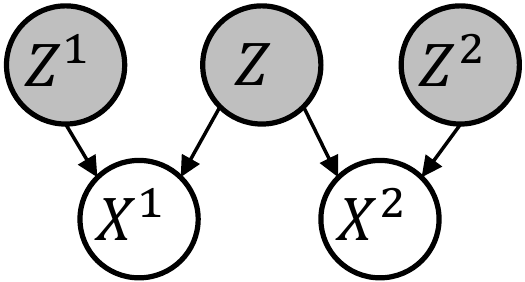}
 \caption{}
 \label{fig:complementary_principle}
\end{subfigure}
\vspace{-1em}
\captionof{figure}{\textbf{\emph{Assumed} compositional structure.}}
\vspace{-1em}
\end{figure}

\textbf{Multi-view learning.} Many existing methods assume that good bits of information co-occur in multiple views, and aim to extract globally present information. Figure \ref{fig:consensus_principle} depicts the generative view of this assumption. One predominant approach is to obtain a \emph{joint representation} by integrating view representations onto the same latent space $z = f\left(g^1(x^1),\ldots,g^k(x^k)\right)$. \citet{ngiam2011multimodal} leveraged stacked autoencoders to obtain joint representation, whereas \citet{srivastava2012learning, srivastava2012multimodal} used probabilistic graphical models to infer $z$. More recent works have used variational autoencoders (VAE) \citep{kingma2013auto}. \citet{suzuki2016joint} introduced a joint encoder structure to learn joint representations, whereas \citet{wu2018multimodal} and \citet{shi2019variational} proposed to combine view representations into a joint representation using product-of-experts (PoE) and mixture-of-experts (MoE) respectively. 

Another approach learns \emph{coordinated representations} by placing regularization $\phi(\cdot)$ on the correlation structure between representations to create a coordinated latent space, i.e.\ $\smash{\argmax_{{h}_{1:K}}\phi({h}_{1:K})}$. Prominent methods are based on canonical correlation analysis (CCA), which learns a common space where the linear canonical correlation between two views is maximized  \citep{hardoon2004canonical}. Subsequent works have introduced non-linear extensions \citep{akaho2006kernel,andrew2013deep,wang2015deep}. These methods rely heavily on pair-wise coordination and cannot efficiently scale to more views. To address this, \citet{benton2017deep} generalized CCA-style analysis to more than two views. Recent works have also adopted \emph{self-supervised learning} (SSL) objectives, which roughly maximize the mutual information between paired views. \citet{federici2019learning} employs a mutual information bottleneck (MIB) to only retain mutual information between views. CLIP \citep{radford2021learning} contrastively maximizes (minimizes) cosine similarity of paired (unpaired) image-text samples. 

The training objectives in CCA and SSL-based methods explicitly encourage learning of a view-invariant representation. A similar effect is implicit in joint representation methods, which can discard localized variations in shared representation spaces \citep{daunhawer2021limitations,wolff2022mixture}. When employed ITW, this bias towards global information can lead to fine-grained localized information being overlooked. Recent methods have additionally sought to preserve view-specific information (the generative model view of this assumption is presented in Figure \ref{fig:complementary_principle}). MFM \citep{tsai2018learning} factorizes $z$ into view-specific factors and shared factors but requires label information. Perhaps most similar to our work, \citet{ye2016learning} and DMVAE \citep{lee2021private} aim for decomposable representations by explicitly separating shared and view-specific factors. However, both methods still rely on assumptions of global information and additionally, target information that manifests privately in each view. Our work does not require compositional assumptions and is capable of learning appropriate aggregation by accounting for inter-view relationships. We compare representative works in \Cref{tab:related_works}.

\textbf{GNN.} Graph autoencoders map graphs into a representation space to subsequently decode graph information from latent representations. \citet{wang2016structural,simonovsky2018graphvae} embeds a graph into a continuous representation $z \in \mathbb{R}^r$ to ensure topologically close nodes have similar representations. \citet{you2018graphrnn} focuses on graph generation, recursively learning node embeddings to generate a graph sequentially. Instead of graph embeddings, \citet{kipf2016variational} infers a latent embedding for each node in the input graph. These works focus on learning embeddings on a fixed input graph $G^{(0)}$, making them unsuitable for our purpose of dynamic and localized information aggregation. Our method is novel in that it hierarchically learns a smaller latent graph $G^{(z)}$ whose node embeddings represent locally aggregated information. We provide an overview of related architectures in \Cref{tab:related_gnns}.

Previous methods have used GNNs for multi-view data by either: \textbf{1.\ }processing each view as a separate graph and using GNN to integrate node representations between graphs \citep{kim2020hypergraph,ma2020multi}; or \textbf{2.\ }constructing an instance graph, where nodes represent instances of the data and edges represent relationships between them across views \citep{wei2019mmgcn,gao2020mgnn}. In this work, we are the first to represent views as nodes and learn edge weights to indicate view dependencies. 

\begin{table}[t!]
\captionof{table}{\textbf{GNN methods.} Overview of representative \emph{graph autoencoder} architectures based on \textbf{encoder/decoder (Enc/Dec)} architecture, \textbf{latent representation (Lat Rep)} and \textbf{aim}. Sim = similarity measure, DP = decision process.}
\label{tab:related_gnns}
\centering
\begin{adjustbox}{max width=0.48\textwidth}
\midsepremove
\begin{tabular}{|p{0.378\columnwidth}|p{0.2\columnwidth}|p{0.162\columnwidth}|p{0.2\columnwidth}|} \toprule
\textbf{Method} & \textbf{Enc/Dec} & \textbf{Lat Rep} & \textbf{Aim} \\ \midrule
\citet{wang2016structural} & MLP/MLP & $z \in \mathbb{R}^r$ & \multirow{2}{*}{\begin{tabular}[c]{@{}l@{}}Node \\ embeddings\end{tabular}} \\ \cmidrule{1-3}
\citet{kipf2016variational} & GNN/Sim & $A, H^{(z)}$ &  \\ \midrule
\citet{you2018graphrnn} & RNN/DP & $A, H^{(z)}$  & \multirow{3}{*}{\begin{tabular}[c]{@{}l@{}} \\Graph \\ generation\end{tabular}} \\ \cmidrule{1-3}
\citet{simonovsky2018graphvae} & GNN/MLP & $z \in \mathbb{R}^r$ &  \\ \cmidrule{1-3}
\citet{de2018molgan} & GNN/MLP & $z \in \mathbb{R}^r$ &  \\ \midrule
\cellcolor[HTML]{C0C0C0}\textbf{LEGATO} & \cellcolor[HTML]{C0C0C0}\textbf{GNN/GNN} & \cellcolor[HTML]{C0C0C0} $\mathbf{A^{(z)}, H^{(z)}}$ &\cellcolor[HTML]{C0C0C0} \begin{tabular}[c]{@{}l@{}}\textbf{Information} \\ \textbf{aggregation} \end{tabular} \\ \bottomrule
\end{tabular}
\midsepdefault
\end{adjustbox}
\vspace{-1em}
\end{table}

\section{Empirical Investigations} \label{sec:experiments}
Having introduced the challenges of learning from multi-view data ITW and our proposed method to address it, we now turn to quantitatively evaluating our method:
\begin{enumerate}[wide,labelwidth=0pt,labelindent=0pt,nolistsep]
    \item \textbf{Learning ITW:} \textit{What is the problem?} \Cref{sec:simulation} employs a simulation of ITW multi-view data to probe the performances of different compositional assumptions.
    \item \textbf{Insights:} \textit{How does it work?} We use  interpretability methods to interpret the graphs and latent aggregations.
    \item \textbf{Performance:} \textit{Does it work?} \Cref{sec:tcga} evaluates downstream performance of our method against state-of-the-art benchmarks on real world dataset.
    \item \textbf{Gains:} \textit{Why does it work?} We deconstruct our method to investigate its sources of performance gain.
\end{enumerate}

\textbf{Benchmarks.} We evaluate our method against 7 state-of-the-art methods, in line with benchmarks found in recent works \citep{federici2019learning,zhang2019ae2,lee2021private}. We consider two coordinated representation methods: \textbf{DCCAE} \citep{wang2015deep} and \textbf{DGCCA} \citep{benton2017deep}; three joint representation methods: \textbf{JMVAE} \citep{suzuki2016joint}, \textbf{MVAE} \citep{wu2018multimodal}, and \textbf{DMVAE} \citep{lee2021private}; and one SSL method: \textbf{MIB} \citep{federici2019learning}. We also include a vanilla \textbf{Transformer} model \citep{vaswani2017attention}, which takes in a sequence of view embeddings and is pretrained using a reconstruction loss. For all results, we report the mean $\pm$ std averaged over $10$ runs. Our implementation can be found at \url{https://github.com/tennisonliu/LEGATO} and at the wider lab repository \url{https://github.com/vanderschaarlab/LEGATO}. We provide additional information about implementation details, dataset preprocessing, and hyperparameters tuning in \Cref{app:implementation_details}.

\subsection{Synthetic Simulation} \label{sec:simulation}
In \Cref{sec:problem_definition}, we characterized real-world ITW data as having more complex view dependencies, giving rise to clusters of localized information, and a larger number of views. In this subsection, we investigate the effect of these two characteristics on the quality of representations. We consider two view correlation settings, $\blacktriangleright \texttt{global}:$ all views are globally correlated with each other, and $\blacktriangleright \texttt{local}:$ views are locally correlated. We construct the following simulation as it is difficult in practice to have natural datasets that possess the required degree of view interaction.

\textbf{Simulation setting.}~We simulate multi-view data with $K<10$ views. Each view is generated from a scalar latent variable such that $\smash{z_k\sim\mathcal{N}(k, 1)}$ and $\smash{z_k \rightarrow x_k}$ $\smash{\:\forall\: k \in [K]}$. We simulate $\texttt{global}$ correlation between views by computing $\smash{z_k \leftarrow (1-w)\cdot z_k + w\cdot z_1 \:\forall\: k \in [K]}$, such that information from $z_1$ is shared across all views. Additionally, $w$ controls how much information is shared, with a larger $w$ indicating higher degrees of overlap, and $w=0$ meaning each view is mutually independent. To simulate $\texttt{local}$ correlation, we sample each pair of latent variables from a multivariate normal distribution, i.e.:
\begin{equation*}
    z_1, z_2 \sim \mathcal{N}(\mathbf{\mu}, \mathbf{\Sigma}), \:\: \textrm{where} \:\: \mathbf{\mu} = \begin{bmatrix} 1 \\ 2\end{bmatrix}, \: \mathbf{\Sigma}=\begin{bmatrix} 1 & w \\ w & 1 \end{bmatrix}
\end{equation*}
which with $K$ views would give us $K/2$ localized clusters, where each cluster of two views is correlated while being mutually independent of other clusters. We generate $100$-dimensional feature vectors for each view using a non-linear transformation, $\smash{x_k=MLP_k(z_k)}$, where $\smash{MLP_k(\cdot)}$ is a randomly initialized single-layer MLP with \texttt{Tanh}$(\cdot)$ activation. The downstream task is the recovery of view-specific latent variables $\smash{\{z\}_{i=1}^K}$, which is a good proxy for whether representations learn localized information. 
\begin{figure}[t!]
     \centering
     \begin{subfigure}[b]{0.9\columnwidth}
         % \centering
         \includegraphics[width=\linewidth]{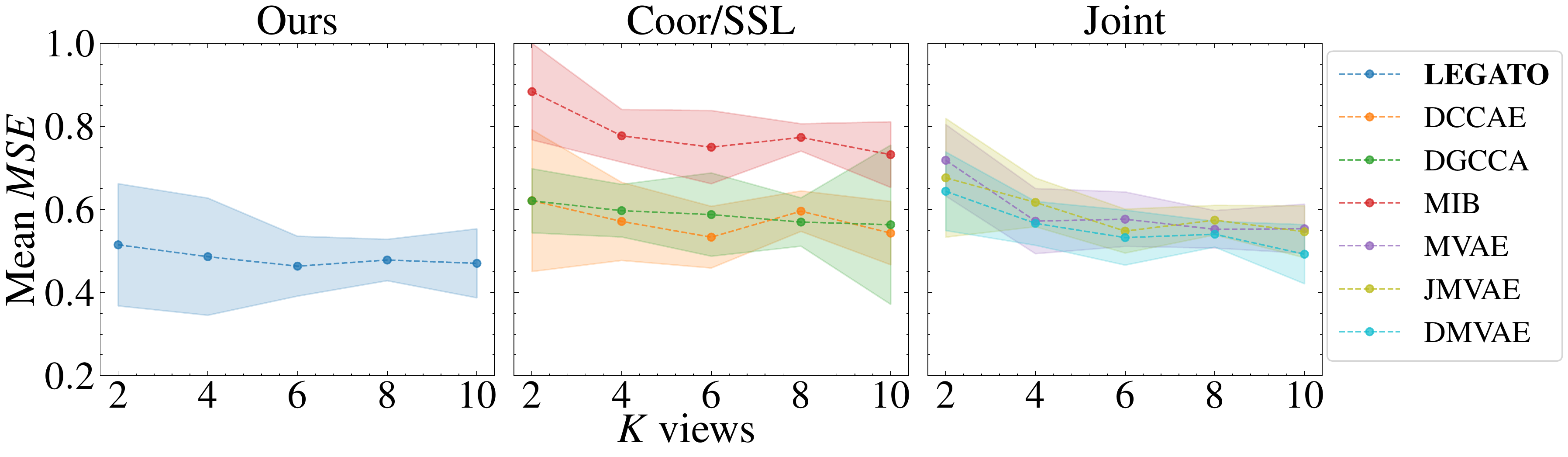}
         \caption{Views are \emph{globally} correlated.}
         \label{fig:global_corr_K}
     \end{subfigure}
     % \hfill
     \begin{subfigure}[b]{0.9\columnwidth}
         % \centering
         \includegraphics[width=\linewidth]{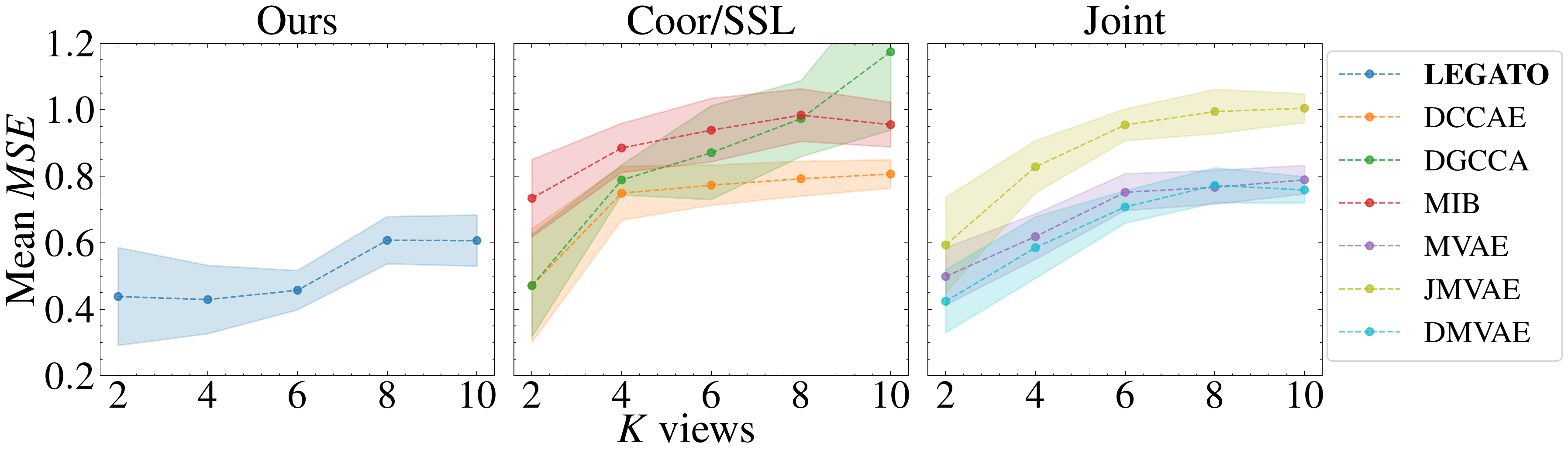}
         \caption{Views are \emph{locally} correlated. }
         \label{fig:local_corr_K}
     \end{subfigure}
\caption{\textbf{Effect of $K$ on learning ($\mathbf{w}$=$\mathbf{0.5}$).} When views are globally correlated, more views lead to better performance. When local correlation increases, performance worsens as more localized clusters of information emerge.}
\label{fig:sim_K}
\vspace{-1em}
\end{figure}
\begin{figure}[h!]
     \centering
     \begin{subfigure}[b]{0.9\linewidth}
         \includegraphics[width=\linewidth]{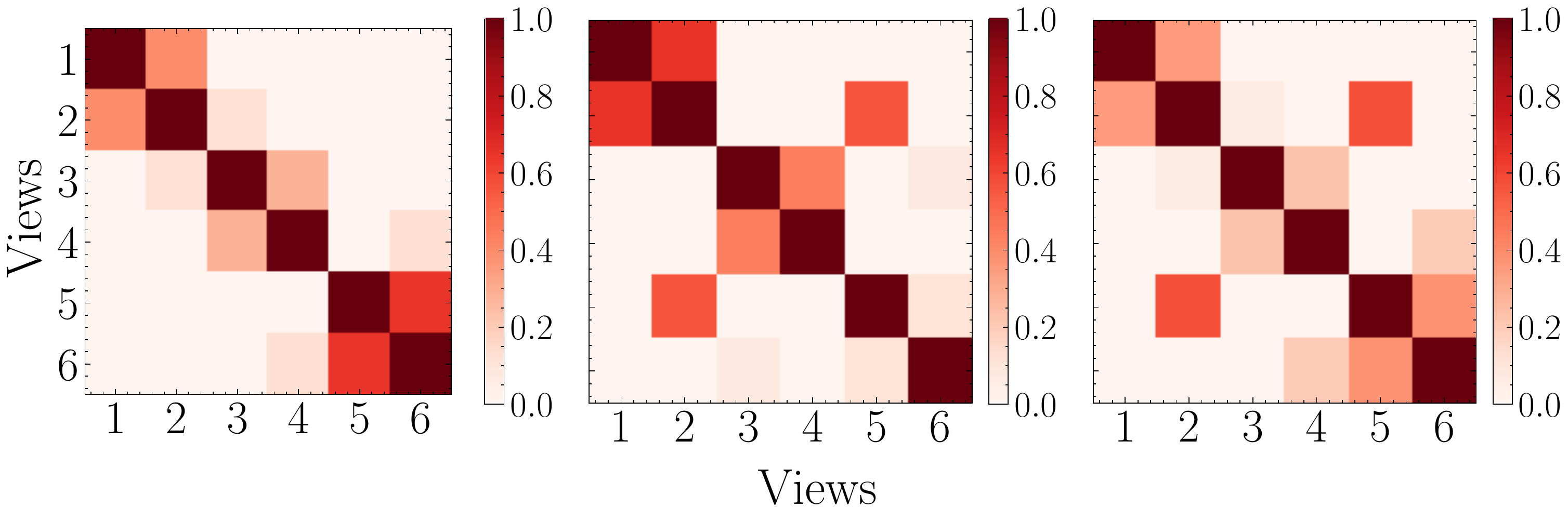}
         \caption{Learned edge weights $\in [0, 1]$ in multi-view graph.}
         \label{fig:learned_adj}
     \end{subfigure}
     \begin{subfigure}[b]{0.9\linewidth}
         \includegraphics[width=\linewidth]{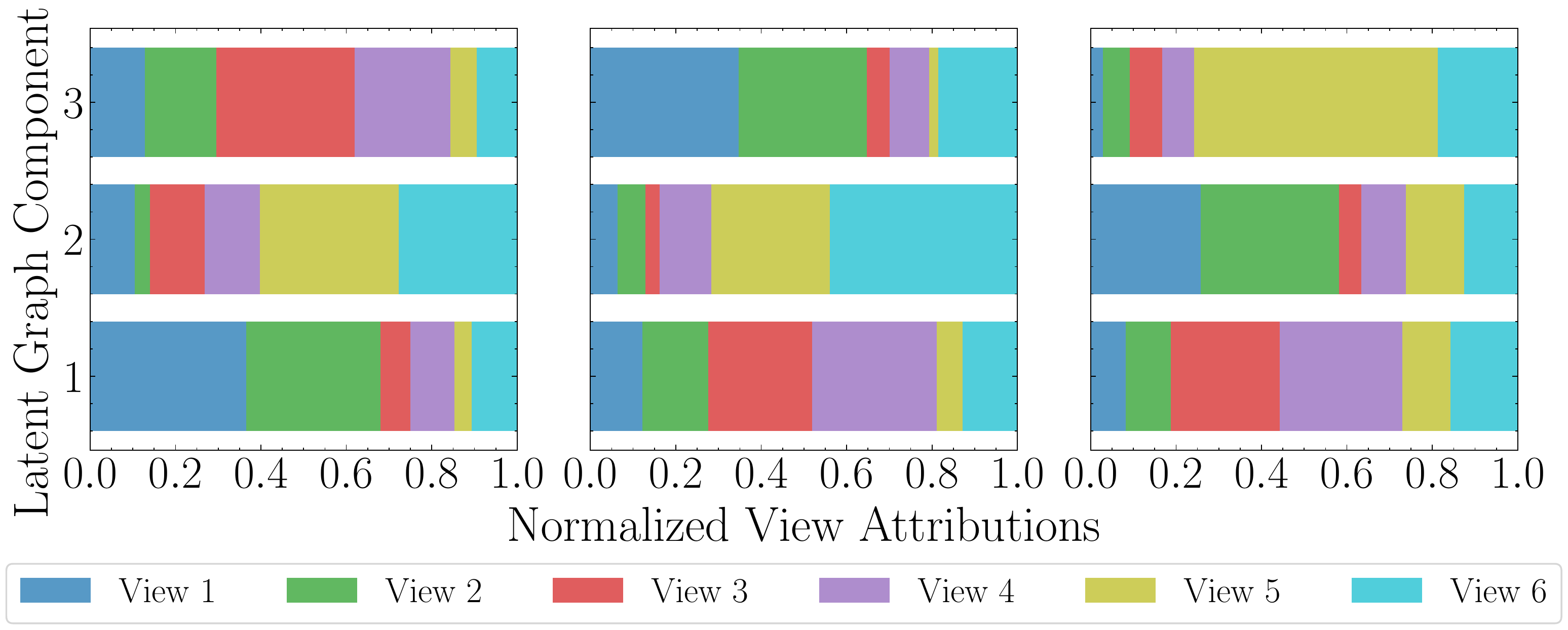}
         \caption{Contribution of views to latent node embeddings.}
         \label{fig:view_contribution}
     \end{subfigure}
\caption{\textbf{Model inspection ($\mathbf{K}$=$\mathbf{6}$).} Our method dynamically learns view dependencies and latent nodes (components) specialize in aggregating localized information.}
\label{fig:interp}
\vspace{-1em}
\end{figure}

\textbf{Results.}~We consider $w$ in range $\{0.0, 0.25, 0.50, 0.75\}$ and $K$ in range $\{2, 4, 6, 8, 10\}$. We plotted the effect of view correlation $w$ and the number of presented views $K$ on representation quality by evaluating the mean $MSE$ in \Cref{fig:failure_mode,fig:sim_K} respectively. As we previously noted, higher global correlation improves the performance of all models, as each view contains more information about all other views. However, increased local correlation is found to decrease the performance of existing methods, which are biased by their compositional assumptions to overlook localized information in favor of globally present factors. Additionally, while views are globally correlated, a larger number of views lead to better performance. In contrast, when views are locally correlated, performances of conventional methods deteriorate quickly as more localized clusters emerge. In comparison, our work is the only one that can effectively learn localized information with higher degrees of local correlation and a larger number of views.

\textbf{Model inspection.}~We investigate the inner workings of our proposed method and the learned multi-view graph and latent graph embeddings. We visualize learned dependencies in the multi-view graphs in \Cref{fig:learned_adj} and use Integrated Gradients \citep{sundararajan2017axiomatic} to visualize the contribution of each view to latent node embeddings in \Cref{fig:view_contribution}. We note that, while our method is not designed for \emph{relational inference}, it can dynamically learn dependencies between views. Additionally, we see the specialization of latent nodes to aggregate information from different regions of the input, where each node focuses on extracting information from more correlated views.

\subsection{Overall Performance} \label{sec:tcga}
\textbf{Datasets.}~We now move on to evaluate our method on three real-world datasets. $\blacktriangleright$ \textbf{TCGA} \citep{tomczak2015review} is a multi-omics dataset containing $7295$ cancer cell lines with $4$ views: mRNA expressions, DNA methylation, microRNA expressions, and reverse-phase protein array. The downstream task is to predict one-year mortality from cancer. $\blacktriangleright$ \textbf{UK Biobank} \citep{sudlow2015uk} is a large population-based medical database. We extract a lung mortality dataset containing $9$ views based on the given feature categorizations.\footnote{\url{https://biobank.ctsu.ox.ac.uk/crystal/cats.cgi}} The views include patient demographics, view and lifestyle factors, physical measures, recorded medical conditions, biomarkers, physical measures, geographical information, treatment history, and family/heredity conditions. The downstream task is the binary classification of lung cancer mortality. $\blacktriangleright$ \textbf{UCI-MFS} \citep{van1998handwritten} is more representative of a traditional multi-view task, where views share similar information. Here, all views contain hand-crafted features extracted from images of handwriting. The downstream task is to predict the handwritten numerals ($0$-$9$). We describe dataset characteristics and pre-processing in \Cref{app:implementation_details}.

\begin{table}[t!]
\centering
\captionof{table}{\textbf{Downstream classification results on three multi-view datasets.} Bold indicates the best performance.}
\label{tab:downstream_results}
\midsepremove
\begin{adjustbox}{max width=1\linewidth}
\begin{tabular}{l|l|c|c|c|} \cmidrule{2-5}
& \textbf{Method} & \begin{tabular}{@{}c@{}}\textbf{TCGA} \\ \textbf{(AUROC $\uparrow$)}\end{tabular} & \begin{tabular}{@{}c@{}}\textbf{UK Biobank} \\ \textbf{(AUROC $\uparrow$)}\end{tabular}  & \begin{tabular}{@{}c@{}}\textbf{UCI-MFS} \\ \textbf{(ACC $\uparrow$)}\end{tabular}  \\ \midrule
\multicolumn{1}{|c|}{} & DCCAE & $0.673 \pm 0.047$ &  $0.624 \pm 0.041$ &  $0.742 \pm 0.034 $ \\
\multicolumn{1}{|c|}{} & DGCCA & $0.620 \pm 0.073$ &  $0.669 \pm 0.058$ &  $0.688 \pm 0.031$ \\
\multicolumn{1}{|c|}{} & JMVAE & $0.695 \pm 0.034$ &  $0.718 \pm 0.043$ &  $0.825 \pm 0.057$ \\
\multicolumn{1}{|c|}{} & MVAE & $0.656 \pm 0.039$ &  $0.715 \pm 0.059$ &  $0.818 \pm 0.042$ \\
\multicolumn{1}{|c|}{} & DMVAE & $0.676 \pm 0.029$ &  $0.688 \pm 0.049$ &  $\mathbf{0.825 \pm 0.043}$ \\
\multicolumn{1}{|c|}{} & MIB & $0.620 \pm 0.083$ & $0.696 \pm 0.067$ &  $0.813 \pm 0.036$\\
\multicolumn{1}{|c|}{\parbox[t]{2mm}{\multirow{-6}{*}{\rotatebox[origin=c]{90}{\emph{Baselines}}}}} & Transformer &  $0.679 \pm 0.080$ &  $0.711 \pm 0.064$ &  $\mathbf{0.825 \pm 0.029}$ \\ \midrule
\multicolumn{1}{|c|}{} & NoHier & $0.652 \pm 0.036$ &  $0.710 \pm 0.041$ &  $0.782 \pm 0.034$ \\
\multicolumn{1}{|c|}{} & NoGraph & $0.696 \pm 0.032$ &  $0.698 \pm 0.030$ &  $0.794 \pm 0.046$ \\
\multicolumn{1}{|c|}{\parbox[t]{2mm}{\multirow{-3}{*}{\rotatebox[origin=c]{90}{\emph{Ablation}}}}} & NoReg & $0.688 \pm 0.039$ &  $0.679 \pm 0.032$ &  $0.801 \pm 0.037$ \\ \midrule
& \cellcolor[HTML]{C0C0C0}\textbf{LEGATO} & \cellcolor[HTML]{C0C0C0}$\mathbf{0.703 \pm 0.051}$ & \cellcolor[HTML]{C0C0C0}  $\mathbf{0.720 \pm 0.038}$ & \cellcolor[HTML]{C0C0C0} $0.824 \pm 0.030$ \\ \cmidrule{2-5}
\end{tabular}
\end{adjustbox}
\midsepdefault
\vspace{-1em}
\end{table}

\textbf{Ablation study.}
Our method is designed with a number of characteristics in mind. Having empirically demonstrated strong overall results, an immediate question is how important these characteristics are for performance. Specifically, we consider the sources of gain from \textbf{(a)} \textit{hierarchical graph pooling} (\textbf{NoHier}), we consider removing the pooling layer, relying simply on GCN layers, \textbf{(b)} \textit{multi-view graph learning} (\textbf{NoGraph}), we replace the learned input graph with a fully-connected graph, and \textbf{(c)} \textit{orthogonality regularization} (\textbf{NoReg}), we remove the orthogonality regularization.

\textbf{Results.}
We report downstream classification performance in \Cref{tab:downstream_results}. We first analyze the performance on \textbf{TCGA} and \textbf{UK Biobank}, which are more representative of tasks found \emph{in-the-wild}, with more complex view dependencies and a higher number of views. We note that in these settings, \texttt{LEGATO} achieves superior performance, being particularly suited for learning the complex dependencies between views and aggregating localized information. We additionally find that joint representation methods perform better than their coordinated counterparts, likely as the emphasis on shared information aggregation is implicit rather than explicitly enforced in CCA and SSL methods. Next, we investigate performance on \textbf{UCI-MFS}, which is more representative of traditional multi-view tasks. Here, we find that our model performs on par with state-of-the-art methods. This is likely because the multi-view assumption holds true, empowering baseline methods (e.g.~\textbf{DMVAE}, \textbf{Transformer}) that exploit the multi-view inductive bias. On our ablation settings, we observe all three aspects are crucial for performance, with a notable $8\%$ performance gain over a GCN network with no latent graph learning. Similarly, orthogonality regularization improves model performance by encouraging orthogonal components. We note that this is more crucial on ITW datasets with more views, as this regularization better encourages the learning of localized information.

\section{Discussion}
Existing multi-view methods make compositional assumptions on the existence of global information, often neglecting localized information when deployed on tabular data ITW. In this work, we represent multi-view data as graphs and their dependencies as learnable edge weights. Moreover, we propose \texttt{LEGATO}, a novel autoencoder that learns a latent graph as a decomposable representation, where each of the latent components specializes in learning different aspects of localized information. Our method empirically demonstrated its effectiveness in learning representations on traditional multi-view tasks but excelled on ITW multi-view datasets with  more complex localized dependencies. \textbf{Future works.} We see several directions for future research. One avenue is the development of better GNN or attention mechanisms tailored to capture localized dependencies more effectively. Additionally, investigating advanced optimization strategies, regularization techniques, and loss functions that account for the specific challenges of multi-view learning in tabular data could lead to improved model performance and generalization. Lastly, while we used an unsupervised reconstruction loss, we believe that the incorporation of more advanced semi- and self-supervised objectives can better leverage unlabeled data to enhance representation learning. 

\section*{Acknowledgements}
We thank the anonymous ICML reviewers as well as members of the van der Schaar lab for many insightful comments and suggestions. Tennison Liu would like to thank AstraZeneca for their sponsorship and support. Jeroen Berrevoets thanks W.D. Armstrong Trust for their support. This work is also supported by the National Science Foundation (NSF, grant number 1722516) and the
Office of Naval Research (ONR).

\bibliography{references}
\bibliographystyle{icml2023}

\newpage
\appendix
\onecolumn
\section{Feature Sets In-The-Wild} \label{app:itw_data}
Multi-view observations contain multiple observations of the same phenomenon and can originate from different modalities (e.g. image and text) but also be multiple observations of the same modality (e.g. multiple tabular datasets). Multi-view learning is the method to integrate information from multiple senses to interact with the world—we see objects, hear sounds, smell odors, and feel texture. Neuroscience research has shown that the brain jointly integrates information from multiple origins and that such synthesis is crucial to reasoning even without explicit labels for multi-view observations \citep{quiroga2009explicit}. While humans can easily learn through multiple senses in an unsupervised way, training a machine with analogous capabilities is a more challenging task. 

The classic multi-view inductive bias hypothesis suggests that views provide the same task-relevant information \citep{yan2021deep}. This assumption is helpful for many problems encountered in image, speech, and text domains. However, this is because data collection procedures in these problems, for example, audio-visual speech recognition \citep{huang2013audio} and image-caption models \citep{radford2021learning}, are carefully controlled to ensure views align and provide the same task-relevant information. However, multi-view data collected in less-controlled, in-the-wild settings are rarely aligned to the same degree. This is especially the case in tabular feature sets, where the higher heterogeneity across feature sets obscures their relationships. Here we consider a few examples:
\begin{itemize}
    \item \textbf{Biobanks.} Examples of this include UK Biobank \citep{bycroft2018uk} and Biobank Japan \citep{nagai2017overview}. These are large-scale biomedical databases that gather a variety of information including physical measures, lifestyle data, cognition and hearing functions, biomarkers, genetic data, and health outcomes. In these precision health datasets, each view provides information on a different aspect of the patient's medical state. It is far more likely for different sources of information to manifest in different clusters of views than for the information to be globally shared across all views.
    \item \textbf{Multi-omics.} Examples of this include The Cancer Genome Atlas Program \citep{tomczak2015review} and Cancer Cell Line Encyclopedia \citep{ghandi2019next}. These problems combine datasets of different omic groups for biological analysis, including genetic, RNA splicing, DNA methylation, histone H3 modification, microRNA expressions, and also subject lineage and ethnicity data, and therapeutics data. Specific disease biomarkers likely manifest in only certain omic groups.
    \item \textbf{Stock market} \citep{ghosh2022forecasting}. The stock market is described from multiple measurements, including trading data of individual stocks, different sources of stock market news (e.g.\ tweets, financial reports), technical indicators, market indices, and wider economic indicators. Evidently, different stocks can be highly dependent on other stocks in the same industry and also industry indicators. 
\end{itemize}
In these settings, learning representations that aggregate information globally across all views will not achieve the desired learning effect. Indeed, we argue that the larger number of view and more complex localized dependencies give rise to localized sources of information that exists in clusters of views.

\section{Graph Neural Networks} \label{app:gnn}
Graph Neural Network (GNN) is a type of deep learning model that can operate on graph-structured data, such as a social network or a molecule. They use neural networks to learn and make predictions on the nodes and edges of the graph. A variety of GNNs have been proposed in recent years, including those employing convolutional networks \citep{defferrard2016convolutional,hamilton2017inductive}, recurrent architectures \citep{li2015gated} and recursive networks \citep{scarselli2008graph}. Most of these approaches can be generalized using the neural \emph{message passing} proposed by \citet{gilmer2017neural}, where node representations are iteratively updated by aggregating features from neighboring nodes.

Neural message-passing algorithms can be mathematically described in the following architecture:
\begin{equation}
    H^{(k)}, A^{(k)} = MP\left(A^{(k-1)}, H^{(k-1)}\right)
\end{equation}
where $H^{(K)}$ are the node embeddings computed after $k$ steps of message passing, $A^{(k)}$ is the adjacency matrix, and $MP(\cdot)$ is some message propagation function. There are many ways to implement the message passing function, but generally using a combination of linear transformations and non-linear activations. One popular model is the graph convolutional network (GCN) \citep{kipf2016semi}, where the node-wise update can be described using:
\begin{equation}
    h^{(k)}_i = f\left(W^{(k)}\sum_{j \in \mathcal{N}(i) \cup {i}}\frac{h^{(k-1)}_j}{\sqrt{\tilde{d}_j^{(k-1)}\tilde{d}_i^{(k-1)}}}\right)
\end{equation}
where $f$ is some non-linear function, and $\mathcal{N}(i)$ is the set of all neighboring nodes of $j$ as indicated in $A^{(k-1)}$. $h^{(k-1)}_j$ is the embedding of the j$^{th}$ node in $H^{(k-1)}$ and $\tilde{d}_j^{(k-1)}$ is the j$^{th}$ row in $\tilde{D}$ where $\tilde{D}=\sum_{j}\tilde{A}^{(k-1)}_{ij}$. Finally, $W^{(k)}$ are the learnable weights of the layer $k$. While GCN applies the same transformation to each node embedding to compose messages, RGCN \citep{schlichtkrull2018modeling} considers different edge types to result in different message transformations:
\begin{equation}
    h^{(k)}_i = f\left(\sum_{r\in\mathcal{R}} \sum_{j\in\mathcal{N}_r(i)}\frac{1}{|\mathcal{N}_r(i)|} W^{(k)}_rh^{(k-1)}_j\right)
\end{equation}
where $\mathcal{R}$ denotes the set of edges types and $W_r^{(k)}$ is the transformation matrix for edge type $r$.

\textbf{Graph autoencoders.} Graph autoencoders are a variant of GNNs that map nodes into a compact latent space and decode graph information from the latent representations. They are mainly used to extract low-dimensional embeddings while preserving a graph's topological information. SDNE \citep{wang2016structural} uses a stacked autoencoder to learn a graph embedding that preserves first and second-order proximity in the graph. VGAE \citep{kipf2016variational} encodes both structural information and node feature information at the same time to learn \emph{node embeddings}. An inner product measure on node embeddings to recover graph structural information, encoding the intuition that nodes that are closely connected in the graph should have similar representations. GraphRNN \citep{you2018graphrnn} follows a similar intuition to encode each latent node recursively for the purposes of dynamic graph generation. \citep{simonovsky2018graphvae} embeds a graph into a single vector representation, which is then used to reconstruct both the adjacency matrix and the node feature matrix. Existing works focus on learning graph embeddings given an input graph, but they are inherently ``flat'' and do not learn hierarchical representations. Perhaps similar to our work, \citet{liu2023goggle} used graph autoencoders to learn a generative model for tabular data but learned a flat graph to model dependencies present in a single feature set.

\section{Implementation Details} \label{app:implementation_details}
\subsection{Training and Hyperparameters}
\textbf{Training.} All models are implemented in PyTorch. The data is split 60-20-20 into an unlabeled training set, labeled training set, and test set respectively, and all reported results are averaged over $10$ runs, where different data splits are sampled for each run. All experiments are run on an NVIDIA Tesla K40C GPU.

\textbf{Hyperparameters.} Models are trained using the Adam \citep{kingma2014adam} with default parameters  $\beta_1=0.9$ and $\beta_2=0.999$. For all experiments, we use batch size of $64$, but tune the learning rate $\eta \in \{0.001, 0.01, 0.1\}$ and weight decay $\in \{0.001, 0.01, 0.1\}$. These and other architecture-specific hyperparameter settings (specific hyperparameters discussed below) are searched using Bayesian Optimization \citep{snoek2012practical} with a search budget of $10$ runs, and where the search objective is the validation set loss. Additionally, we employ early stopping to terminate model training after $20$ epochs of no improvement on the validation set, after which the best model is returned for evaluation.

\subsection{Model Implementation}
While we are agnostic to the specific GNN architecture employed in our \texttt{POOL} and \texttt{UNPOOL} layers, we implemented GCN \citep{kipf2016semi} and RGCN \citep{schlichtkrull2018modeling}. Specifically, for RGCN, we employ the basis decomposition proposed in \citet{schlichtkrull2018modeling}:
\begin{equation}
    W^{(k)}_r=\sum_{b=1}^Ba^{k}_rbV_b^{(k)}
\end{equation}
Therefore, where the weights $W^{(k)}_r$ form a weighted combination of a basis transformation $V_b^{(k)}$ with coefficients $a^{(k)}_{rb}$. We choose $B=5$, reducing the number of learnable parameters in our model.

The dimensionality of our latent graph is chosen to be $K'=K/2$, so the latent graph has half the number of nodes as the multi-view graph. We found this to be a robust setting that worked well in our experiments. Additionally, we add a node normalization layer after each layer \citep{ba2016layer,zhou2021understanding}. This reduces the problem of node (or view) dominance when computing new node embeddings, allowing for comparable contributions to the weighted combination in \Cref{eq:refinement_x}. This operation is formally expressed as: $\smash{\text{Norm}(h^k) = (h^k-\mu^k)/\sqrt{var^k}}$, where $\smash{\mu^k}$ and $\smash{var^k}$ are the mean and variance of $h^k$ calculated per dimension over the mini-batch. Lastly, we consider $\alpha\in\{0.001, 0.01, 0.1\}$ and $\beta\in\{0.001, 0.01, 0.1\}$ for our unsupervised learning objective.

\textbf{Encoder/decoder networks.} For both our model and the baselines we compared against, we use view encoders/decoders with a single \texttt{ReLU}-activated hidden layer. We tune the dimensionality of the hidden representation by considering $d\in\{50, 60, 70, 80, 90, 100\}$. We use the same encoder/decoder architectures to ensure a fair comparison.

\subsection{Baseline Implementation}
In this subsection, we provide further details on the implementation of benchmarks we compare against, including \textbf{DCCAE} \citep{wang2015deep}, \textbf{DGCCA} \citep{benton2017deep}, \textbf{JMVAE} \citep{suzuki2016joint}, \textbf{MVAE} \citep{wu2018multimodal}, \textbf{DMVAE} \citep{lee2021private} and \textbf{MIB} \citep{federici2019learning}.

\textbf{DCCAE \citep{wang2015deep}} is trained using two objectives: CCA objective to encourage view embeddings to be similar, and a reconstruction objective, where $\lambda$ is a weighting parameter used to trade off the two objectives. We tune $\lambda \in \{0, 0.5, 1\}$ and use $\varepsilon=0.001$, which is the default setting used to regularize CCA calculations. As \textbf{DCCAE} is designed with two views in mind, we modify the CCA objective when we have more views $\sum_{i=1}^K\phi(h_i, h_1)$. We use the implementation by \citet{Chapman2021}, which is publicly available at \url{https://github.com/jameschapman19/cca_zoo}.

\textbf{DGCCA \citep{benton2017deep}} generalizes CCA objectives to more than two views. We similarly use $\varepsilon=0.001$, tune $\lambda \in \{0, 0.5, 1\}$, and use the implementation by \citet{Chapman2021}.

\textbf{JMVAE \citep{suzuki2016joint}} integrates view embeddings using a neural network to infer a stochastic latent variable $z$. The latent variable is stochastic, and the model is trained using the ELBO loss. We use the implementation available at \url{https://github.com/masa-su/jmvae}.

\textbf{MVAE \citep{wu2018multimodal}} integrates view embeddings using a product-of-expert (POE) model, $q(z|x) = \Pi_{k=1}^Kq(h^k|x)$. We use the implementation available at \url{https://github.com/mhw32/multimodal-vae-public}.

\textbf{DMVAE \citep{lee2021private}} uses a VAE architecture and introduces a separate latent variable $\{z^i\}_{i=1}^K$ for each view. We use the publicly available implementation \url{https://github.com/seqam-lab/DMVAE}.

\textbf{MIB \citep{federici2019learning}} introduces a variational information bottleneck to discard information that is not shared between views, where a hyperparameter $\lambda$ is introduced to bottleneck superfluous information. We consider $\lambda\in \{0, 0.5, 1\}$ and use the implementation at \url{https://github.com/mfederici/Multi-View-Information-Bottleneck}.

\subsection{TCGA Preprocessing} \label{app:tcga}
We analyze 1-year mortality based on the comprehensive observations from multiple omics on $7295$ cancer cell lines (i.e. samples) data consists of observations from $4$ distinct views on each cell line across 3 different omics layers: 1.\ mRNA expressions, 2.\ DNA methylation, 3.\ microRNA expressions, and 4.\ reverse phase protein array.

For constructing multiple views and labels, the following datasets were downloaded from \url{http://gdac.broadinstitute.org}:
\begin{itemize}
    \item DNA methylation (epigenomics): \textit{Methylation\_Preprocess.Level\_3.2016012800.0.0.tar.gz}
    \item microRNA expression (transcriptomics): \textit{miRseq\_Preprocess.Level\_3.2016012800.0.0.tar.gz}
    \item mRNA expression (transcriptomics): \textit{mRNAseq\_Preprocess.Level\_3.2016012800.0.0.tar.gz}
    \item RPPA (proteomics): \textit{RPPA\_AnnotateWithGene.Level\_3.2016012800.0.0.tar.gz}
    \item clinical labels: \textit{Clinical\_Pick\_Tier1.Level\_4.2016012800.0.0.tar.gz}
\end{itemize}

Time to death or censoring in clinical labels was converted to a binary label for 1-year mortality. We imputed missing values within the observed views with mean values. To focus our experiments on the integrative analysis and to avoid \emph{curse-of-dimensionality} in the high-dimensional multi-omics data, we extracted low-dimensional representations (i.e., 100 features) using the kernel-PCA (with polynomial kernels) on each view \citep{shiokawa2018application}.

\subsection{UK Biobank Preprocessing} \label{app:biobank}
We used data from the UK Biobank \citep{sudlow2015uk}, a large prospective cohort of half a million men and women recruited between 2006-10 from across the UK with ongoing follow-up. As lung cancer screening is only considered in ever-smokers, we include individuals without a previous diagnosis of lung cancer at baseline who self-reported as current or former smokers. Lung cancer diagnosis were determined through linked national cancer registry, right censored at 31/07/2019 \citep{sudlow2015uk}.

The lung cancer dataset is extracted from UK Biobank using the scripts provided in \url{https://github.com/callta/synthetic-data-analyses/tree/main/code} by executing preprocessing scripts sequentially. Additionally, we dropped all variables with more than $25\%$ missingness and all rows with more than $1\%$ missingness. We normalized continuous variables such their values lay between 0 and categorical variables were one-hot encoded. To manage missing data, we used mean imputation. Lastly, we extracted relevant $9$ views by using the feature categorizations provided at \url{https://biobank.ctsu.ox.ac.uk/crystal/cats.cgi}. The specific variable names included in each view can be found in the $\texttt{.json}$ files and the preprocessing instructions included in our code at \url{https://github.com/tennisonliu/LEGATO/tree/master/exps/biobank_exp}.

\end{document}